\documentclass{article}
\usepackage{spconf,amsmath,graphicx,hyperref}

\usepackage{graphicx,verbatim}
\usepackage{amsfonts}
\usepackage{amssymb}

\usepackage{xcolor} 
\usepackage{colortbl}
\usepackage{booktabs}
\usepackage{soul}

\name{Yu Li$^{1}$ \qquad Da Chang$^{2}$ \qquad Xi Xiao$^{3}$}
\address{
$^{1}$ The George Washington University, Washington, DC, USA \\
$^{2}$ Shenzhen Institute of Advanced Technology, Chinese Academy of Sciences, Shenzhen, China \\
$^{3}$ University of Alabama at Birmingham, Birmingham, AL, USA \\
}

\begin{document}
\ninept
\title{KG-SAM: Encoding Structural Constraints into Segment Anything Models via Probabilistic Graphical Models}
\maketitle

\begin{abstract}
While the Segment Anything Model (SAM) has achieved remarkable success in image segmentation, its direct application to medical imaging remains hindered by fundamental challenges, including ambiguous boundaries, insufficient modeling of anatomical relationships, and the absence of uncertainty quantification.
To address these limitations, we introduce \textbf{KG-SAM}, a knowledge-guided framework that synergistically integrates anatomical priors with boundary refinement and uncertainty estimation.
Specifically, KG-SAM incorporates (i) a medical knowledge graph to encode fine-grained anatomical relationships, (ii) an energy-based Conditional Random Field (CRF) to enforce anatomically consistent predictions, and (iii) an uncertainty-aware fusion module to enhance reliability in high-stakes clinical scenarios. Extensive experiments across multi-center medical datasets demonstrate the effectiveness of our approach: KG-SAM achieves an average Dice score of 82.69\% on prostate segmentation and delivers substantial gains in abdominal segmentation, reaching 78.05\% on MRI and 79.68\% on CT. These results establish KG-SAM as a robust and generalizable framework for advancing medical image segmentation.
\end{abstract}

\section{Introduction}
\label{sec:intro}

Medical image segmentation is a fundamental task in computer vision and clinical practice. Accurate segmentation is essential for diagnosis, treatment planning, and disease monitoring. However, robust performance in real-world clinical scenarios remains challenging due to large variations across patients, imaging modalities, and acquisition protocols. Although deep learning methods have achieved strong results~\cite{gao2023anatomy}, their deployment is often hindered by poor generalization and unstable performance across diverse domains. The recent advent of foundation models, particularly the Segment Anything Model (SAM)~\cite{Kirillov2023SegmentA}, has introduced a new paradigm with its exceptional generalization capabilities, offering immense potential for medical image analysis~\cite{li2024polyp}.

Despite this promise, the direct application of SAM to medical imaging is fraught with fundamental limitations. First, its prompt-driven architecture struggles with the low-contrast and ambiguous boundaries characteristic of many medical scans. Second, SAM operates without an intrinsic understanding of anatomical topology, frequently leading to segmentations that are structurally inconsistent or biologically implausible. Third, its lack of uncertainty quantification poses significant reliability concerns in high-stakes clinical decision-making.
Although recent adaptations involving prompt engineering~\cite{ma2024segment} or fine-tuning with shape priors~\cite{li2025sfmdiffusion} show promise, they often fail to enforce anatomical correctness. Similarly, existing uncertainty quantification techniques~\cite{Zhang2021UncertaintyGuidedMC} typically operate in isolation, neglecting crucial anatomical constraints.

To address these interconnected challenges, we propose KG-SAM, a \textbf{K}nowledge-\textbf{G}uided framework that simultaneously enforces boundary refinement, anatomical consistency, and uncertainty awareness. The cornerstone of our framework is a medical knowledge graph that encodes explicit anatomical relationships and spatial constraints. The anatomical knowledge is then integrated with visual features via an energy-based Conditional Random Field (CRF) optimization module. The CRF's energy function is designed to jointly minimize boundary ambiguity while penalizing anatomically inconsistent predictions. The framework is further enhanced by a probabilistic fusion mechanism that intelligently weighs segmentation outputs against their uncertainty estimates, bolstering the final result's reliability. Through extensive validation on multi-center cardiac and prostate imaging datasets, KG-SAM achieves state-of-the-art performance with an average Dice score of 82.69\%, demonstrating a 14.71\% improvement over baseline SAM implementations in challenging segmentation tasks. Our ablation studies reveal that the synergistic combination of components contributes a 10.60\% performance gain. Furthermore, the orthogonal nature of our method is demonstrated through its integration with DeSAM~\cite{gao2024desam}, where KG-De-SAM achieves an enhanced Dice score of 84.52\%, showing an additional 2.27\% improvement over DeSAM. Our main contributions are as follows:
\begin{itemize}
    \item We introduce \textbf{KG-SAM}, a knowledge-guided extension of SAM that unifies anatomical priors, boundary refinement, and uncertainty quantification for medical image segmentation.
    \item We design an energy-based CRF module that integrates a medical knowledge graph with SAM features, enforcing anatomical consistency and reducing boundary ambiguity.
    \item We demonstrate state-of-the-art results on multi-center datasets and show that KG-SAM is modular, seamlessly boosting existing SAM variants such as DeSAM with additional performance gains.
\end{itemize}

\begin{figure*}[htbp]
\centering
\includegraphics[width=1.85\columnwidth]{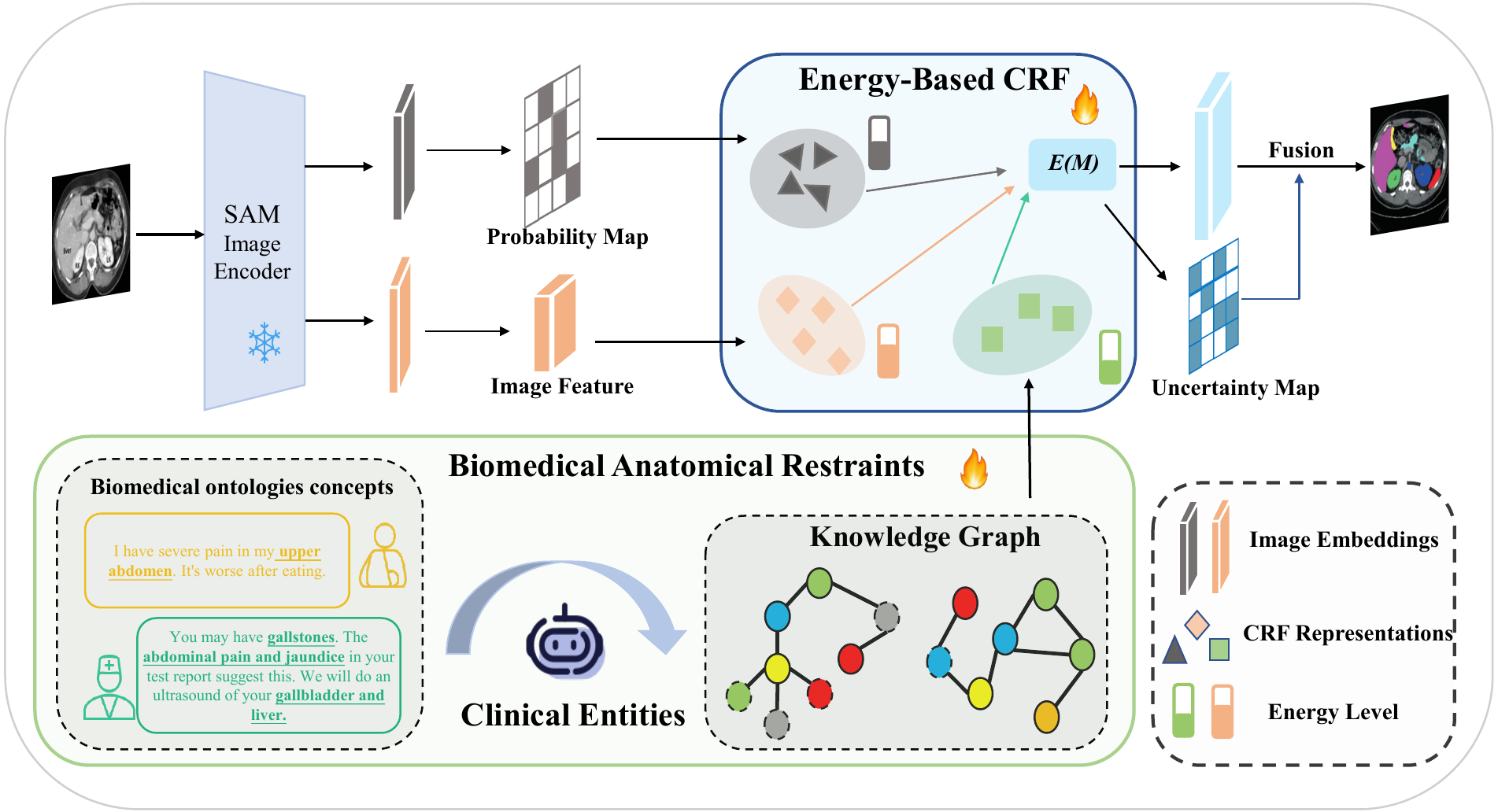}
\caption{An overview of our proposed KG-SAM framework. A frozen SAM image encoder extracts deep image embeddings and an initial probability map from the input image. A knowledge graph is constructed from biomedical concepts to explicitly model anatomical priors. The core of the framework is the Energy-Based CRF module, which integrates the visual features from SAM with the anatomical constraints from the knowledge graph. Finally, the Fusion module uses the uncertainty map to adaptively weight features and produce the final, anatomically coherent segmentation. The components for CRF optimization and fusion are learnable.}
\label{fig:model}
\end{figure*}

\vspace{-2pt}
\section{Related Work}

\subsection{SAM and Uncertainty in Medical Segmentation}
The adaptation of the Segment Anything Model (SAM) for medical imaging has rapidly evolved. Initial approaches focused on fine-tuning to specialize the model for specific domains~\cite{ma2024segment}. Subsequent work explored more advanced strategies, including prompt engineering to better guide the model's focus~\cite{zhang2024segment,lei2025medlsam} and sophisticated architectural modifications, such as dual-branch fusion networks, to improve performance on specific tasks like breast ultrasound segmentation~\cite{li2025dual}. However, a fundamental limitation underlies these adaptations: they generally lack an explicit model of anatomical topology, often resulting in segmentations that are locally plausible but globally inconsistent. Beyond anatomical implausibility, a second critical hurdle for clinical deployment is the quantification of predictive uncertainty. While established techniques like Bayesian neural networks and ensemble methods exist for this purpose~\cite{mehrtash2020confidence}, however, most existing methods treat uncertainty estimation as a standalone component, without effectively integrating it with anatomical constraints or boundary optimization.

\subsection{Conditional Random Fields in Image Segmentation}
Conditional Random Fields have a long and successful history in image segmentation, valued for their ability to impose structural consistency on pixel-wise predictions. Early applications primarily utilized pairwise potentials to encourage local smoothnes, while later work incorporated higher-order potentials for more complex relationships~\cite{xu2022image}. The modern paradigm integrates CRFs with deep neural networks~\cite{wang2022conditional}, often as a final refinement layer that can be optimized end-to-end.
Nevertheless, the priors in most deep CRF models are often limited to generic, local smoothness constraints. They typically lack a mechanism to incorporate explicit, long-range anatomical knowledge, such as the relative positioning or size constraints between organs, which limits their efficacy in complex medical imaging scenarios where precise structural understanding is crucial.

\vspace{-2pt}
\section{Methodology}
\subsection{Overview}
As illustrated in Figure~\ref{fig:model}, our KG-SAM framework is designed as a two-stream architecture that synergizes a vision foundation model with explicit medical knowledge. The process begins with the SAM Image Encoder, which takes a medical image as input to produce an initial probability map and a set of deep image features. Concurrently, the knowledge stream constructs a Knowledge Graph by extracting clinical entities from biomedical concepts to represent anatomical priors. The core of our framework is the Energy-Based CRF module, which receives the outputs from both streams. It optimizes an energy function, $E(M)$, to integrate the anatomical constraints from the knowledge graph with the visual evidence from SAM. This process yields a refined segmentation and a corresponding Uncertainty Map. Finally, a Fusion step utilizes the uncertainty map to produce the final, anatomically coherent segmentation result.

\subsection{Image Embedding Extraction from SAM}
Our feature extractor is the Vision Transformer (ViT) encoder from the Segment Anything Model (SAM). An input medical image $\mathbf{I} \in \mathbb{R}^{H \times W}$ is preprocessed by resizing to $1024 \times 1024$ and normalizing pixel values, resulting in an input tensor $\mathbf{I}_{\text{in}} \in \mathbb{R}^{1024 \times 1024 \times 3}$. The ViT encoder, $f_{\text{enc}}$, processes $\mathbf{I}_{\text{in}}$ to generate hierarchical features. We extract multi-scale embeddings $\{\mathbf{E}_l\}_{l \in L}$ from intermediate layers of the encoder, where each $\mathbf{E}_l \in \mathbb{R}^{H_l \times W_l \times C_l}$ represents features at a specific scale. The extracted embeddings are then processed by a decoder head composed of residual blocks to produce a refined feature map $\mathbf{F}_{\text{ref}}$. This map is subsequently fed into two parallel branches:
\begin{enumerate}
    \item \textbf{Segmentation Branch:} This branch generates an initial probability map $\mathbf{P} \in [0, 1]^{H \times W}$ using convolution and upsampling. This map provides the unary potentials for the subsequent CRF optimization.
    \item \textbf{Feature Branch:} This branch outputs a dense feature representation $\mathbf{F}_{\text{feat}} \in \mathbb{R}^{H' \times W' \times D}$, which is used to compute the pairwise potentials within the CRF.
\end{enumerate}

\subsection{Knowledge Graph Construction}
To encode anatomical priors, we construct a medical knowledge graph defined as $\mathcal{G} = (\mathcal{V}, \mathcal{E}, \mathcal{R})$. Its vertex set $\mathcal{V}$ consists of nodes $\{v_i\}$ representing anatomical structures, each associated with a feature vector $\mathbf{h}_i$ encoding its intrinsic properties. The edge set $\mathcal{E}$ captures relationships between nodes, such as spatial adjacency and hierarchical subdivisions, where each edge $e_{ij}$ is characterized by a relationship type $r_{ij}$ and a weight $w_{ij}$ learned from medical data. The rule set $\mathcal{R}$ incorporates domain-specific constraints derived from sources like radiology reports and anatomical atlases, which are formalized into an adjacency constraint matrix $\mathbf{A}$. To align the knowledge graph with the image space, we learn an affine transformation $\mathbf{T}_{\text{affine}}$ that maps predefined image landmarks to a standardized anatomical atlas. This registration process ensures that the spatial reasoning based on node features $\{\mathbf{h}_i\}$ is consistent across different patient images.

\subsection{Energy-Based Conditional Random Field Optimization}
We integrate the anatomical constraints from the knowledge graph with the image-based features using an energy-based Conditional Random Field (CRF). The goal is to find the optimal segmentation map $\mathbf{M}$ that minimizes the following energy function:
\begin{equation}
E(\mathbf{M}) = \sum_{i} \psi_u(m_i) + \sum_{i<j} \psi_p(m_i, m_j) + \sum_{(o_1,o_2) \in \mathcal{G}} \psi_a(m_{o_1}, m_{o_2})
\end{equation}
where $m_i$ is the label assigned to pixel $i$, and $m_{o_k}$ refers to the segmentation of organ $o_k$. The energy function consists of three terms:

The unary potential $\psi_u(m_i) = -\log(P(m_i))$ is derived from the initial probability map $\mathbf{P}$ generated by the segmentation branch, where $P(m_i)$ is the probability of pixel $i$ having label $m_i$.

The pairwise potential $\psi_p(m_i, m_j)$ encourages label consistency between adjacent pixels. It is typically modeled as a Gaussian kernel on the feature space: $\psi_p(m_i, m_j) = \mu(m_i, m_j) \exp(-\frac{\|\mathbf{f}_i - \mathbf{f}_j\|^2}{2\sigma^2})$, where $\mathbf{f}_i$ and $\mathbf{f}_j$ are feature vectors from $\mathbf{F}_{\text{feat}}$ for pixels $i$ and $j$, and $\mu$ is a label compatibility function.

The anatomical potential $\psi_a(m_{o_1}, m_{o_2})$ is our novel term that enforces high-level structural constraints from the knowledge graph $\mathcal{G}$. It is defined as:
\begin{equation}
\psi_a(m_{o_1}, m_{o_2}) = w_{o_1,o_2} \cdot \mathcal{L}_{\text{IoU}}(R(m_{o_1}), \mathcal{A}(o_1|o_2))
\end{equation}
where $R(m_{o_1})$ is the predicted region for organ $o_1$, $\mathcal{A}(o_1|o_2)$ is the expected spatial distribution of organ $o_1$ relative to $o_2$ as defined by the knowledge graph, $w_{o_1,o_2}$ is the corresponding edge weight, and $\mathcal{L}_{\text{IoU}}$ is a loss function based on the Intersection over Union metric.

We use mean-field approximation for optimization, which iteratively updates the probability distribution $Q(\mathbf{M})$ to approximate the true posterior. The update rule for the distribution at each pixel $i$ at iteration $t+1$ is:
\begin{equation}
\begin{split}
    Q^{(t+1)}(m_i) = \frac{1}{Z_i} \exp\Biggl(&-\psi_u(m_i) - \sum_{j \neq i} \mathbb{E}_{Q^{(t)}}[\psi_p(m_i,m_j)] \\
    & - \sum_{(o_1,o_2)\in \mathcal{G}} \mathbb{E}_{Q^{(t)}}[\psi_a(m_{o_1},m_{o_2})]\Biggr)
\end{split}
\end{equation}
where $Z_i$ is the partition function. The process continues until the change in the distribution is below a threshold $\epsilon$.

Finally, we compute a pixel-wise uncertainty map $\mathbf{U} \in \mathbb{R}^{H \times W}$. This map combines feature-level ambiguity (aleatoric uncertainty) with the degree of anatomical constraint violation (structural uncertainty):
\begin{equation}
\mathbf{U} = \mathbb{H}(\{\mathbf{P}_m\}_{m=1}^M) + \lambda_a \sum_{(o_1,o_2)\in \mathcal{G}} \| \mathbf{A}_{o_1,o_2} - \mathbf{Q}_{o_1,o_2}^* \|_F
\end{equation}
Here, the first term $\mathbb{H}(\cdot)$ is the entropy computed over $M$ stochastic forward passes (e.g., using Monte Carlo dropout) to estimate model uncertainty from the probability maps $\{\mathbf{P}_m\}$. The second term measures the violation of anatomical rules, where $\mathbf{A}_{o_1,o_2}$ is the constraint from the knowledge graph and $\mathbf{Q}_{o_1,o_2}^*$ is the final optimized pairwise probability between the organs, with $\lambda_a$ as a balancing coefficient.

The final feature map, $\mathbf{F}_{\text{fused}}$, is computed by adaptively weighting multi-level features $\{\mathbf{F}^*_l\}_{l=1}^L$ based on the uncertainty map $\mathbf{U}$. The fusion is defined as:
\begin{equation}
\mathbf{F}_{\text{fused}} = \sum_{l=1}^L \alpha_l(\mathbf{U}) \cdot \mathbf{F}^*_l
\end{equation}
where the weights $\alpha_l(\mathbf{U})$ are calculated from the uncertainty map $\mathbf{U}_l$ for each feature level $l$ via a softmax function:
\begin{equation}
\alpha_l(\mathbf{U}) = \frac{\exp(-\beta \cdot \mathbf{U}_l)}{\sum_{k=1}^L \exp(-\beta \cdot \mathbf{U}_k)}
\end{equation}
The hyperparameter $\beta >$ controls the sharpness of the weighting distribution. The final segmentation is then generated from $\mathbf{F}_{\text{fused}}$.

\begin{table*}[t]
\caption{Comparison with state-of-the-art methods on Abdominal (cross-modality) and Prostate (multi-site) datasets. Dice score (\%) is reported. The best result is in \textbf{bold}, the second-best is \ul{underlined}, and our methods are highlighted.}
\centering
\small
\renewcommand{\arraystretch}{0.8} 
\setlength{\tabcolsep}{2pt}       
\begin{tabular}{lccccccccc}
\toprule
\textbf{Method} & \multicolumn{2}{c}{\textbf{Abdominal}} & \multicolumn{7}{c}{\textbf{Prostate}} \\
\cmidrule(lr){2-3} \cmidrule(lr){4-10}
& MRI & CT & Overall & Site A & Site B & Site C & Site D & Site E & Site F \\
\midrule
\multicolumn{10}{l}{\textit{General Segmentation Methods}} \\
nnU-Net~\cite{isensee2018nnu} & 66.04 & 68.92 & 48.75 & 63.73 & 61.21 & 27.41 & 34.36 & 44.10 & 61.70 \\
\midrule
\multicolumn{10}{l}{\textit{Domain Generalization Methods}} \\
AdvBias~\cite{chen2020realistic} & 71.35 & 72.18 & 60.45 & 77.45 & 62.12 & 51.09 & 70.20 & 51.12 & 50.69 \\
RandConv~\cite{xu2020robust} & 70.57 & 73.85 & 57.07 & 75.52 & 57.23 & 44.21 & 61.27 & 49.98 & 54.21 \\
MaxStyle~\cite{chen2022maxstyle} & 73.81 & 77.96 & 68.27 & 81.25 & 70.27 & 62.09 & 58.18 & 70.04 & 67.77 \\
CSDG~\cite{ouyang2022causality} & 74.62 & 75.43 & 70.06 & 80.72 & 68.00 & 59.78 & 72.40 & 68.67 & 70.78 \\
\midrule
\multicolumn{10}{l}{\textit{SAM-based Methods}} \\
MedSAM~\cite{ma2024segment} & 70.23 & 75.48 & 66.98 & 72.32 & 73.31 & 61.53 & 64.46 & 68.89 & 61.39 \\
SAMed~\cite{zhang2023customized} & 67.42 & 74.36 & 70.23 & 73.61 & 75.89 & 58.61 & 73.91 & 66.52 & 72.85 \\
SAMUS~\cite{lin2023samus} & 68.57 & 74.07 & 76.59 & 76.30 & 77.06 & 75.65 & 76.87 & 77.58 & 76.05 \\
DeSAM~\cite{gao2024desam} & \ul{75.57} & \ul{79.87} & 82.25 & 82.30 & \ul{82.06} & \ul{85.65} & 82.87 & 80.58 & 81.05 \\
\rowcolor{gray!15}
\textbf{KG-SAM (Ours)} & 78.05 & 79.68 & \ul{82.69} & \ul{82.80} & 80.61 & 84.77 & \ul{83.41} & \ul{82.36} & \ul{82.17} \\
\rowcolor{gray!30}
\textbf{KG-De-SAM (Ours)} & \textbf{80.15} & \textbf{81.68} & \textbf{84.52} & \textbf{84.80} & \textbf{83.61} & \textbf{86.77} & \textbf{85.41} & \textbf{83.36} & \textbf{83.17} \\
\bottomrule
\end{tabular}
\label{tab1}
\end{table*}
\vspace{-6pt}

\section{Experimental Results}
\subsection{Dataset and Implementation Details}
We evaluate our method in both cross-domain and single-domain settings using multiple public datasets. For cross-domain evaluation, we use: 1) BTCV dataset~\cite{landman2015miccai} with 13 annotated abdominal organs, 2) AMOS dataset~\cite{ji2022amos} containing 15 organs, and 3) multi-site prostate dataset collected from three public sources including NCI-ISBI-2013~\cite{bloch2015nci}, I2CVB~\cite{lemaitre2015computer}, and PROMISE12~\cite{litjens2014evaluation}.
We followed the standard preprocessing protocol ~\cite{chen2022maxstyle}, resampling images to dataset-specific target spacings and normalizing them to [-1, 1].
Empirically set CRF optimization parameters were $\lambda_f=0.5$, $\beta=1.0$, and $\lambda=0.3$.

\begin{table}[t]
\caption{Ablation study of the core components in KG-SAM. Dice score (\%) is reported for abdominal and prostate segmentation. Baseline is SAM with our custom decoder head. Best results are in \textbf{bold}.}
\centering
\small
\renewcommand{\arraystretch}{0.7} 
\setlength{\tabcolsep}{2pt}       
\begin{tabular}{lcccc}
\toprule
\textbf{Method} & \multicolumn{2}{c}{\textbf{Abdominal}} & \multicolumn{2}{c}{\textbf{Prostate}} \\
\cmidrule(lr){2-3} \cmidrule(lr){4-5}
 & CT & MRI & Overall & Sites (Mean $\pm$ Std) \\
\midrule
Baseline (SAM)      & 71.52 & 70.21 & 72.09 & 72.09 $\pm$ 0.82 \\
+ CRF only          & 73.81 & 72.95 & 76.86 & 76.88 $\pm$ 1.25 \\
+ KG only           & 74.13 & 73.14 & 77.00 & 77.00 $\pm$ 1.16 \\
+ UF only           & 72.68 & 71.05 & 72.68 & 72.69 $\pm$ 1.11 \\
\midrule
Full w/o UF         & 78.53 & 77.14 & 79.86 & 79.87 $\pm$ 1.15 \\
Full w/o KG         & 77.92 & 76.76 & 78.57 & 78.57 $\pm$ 1.19 \\
Full w/o CRF        & 77.89 & 76.54 & 79.13 & 79.13 $\pm$ 1.17 \\
\midrule
\rowcolor{gray!20}
\textbf{Full (Ours)} & \textbf{78.05} & \textbf{79.68} & \textbf{82.69} & \textbf{82.69} $\pm$ \textbf{1.45} \\
\bottomrule
\end{tabular}
\label{tab:ablation}
\end{table}

\subsection{Performance and Comparison}
We benchmark KG-SAM against a comprehensive set of state-of-the-art methods. These include the widely-used nnU-Net~\cite{isensee2018nnu}, several recent domain generalization techniques (AdvBias~\cite{chen2020realistic}, RandConv~\cite{xu2020robust}, MaxStyle~\cite{chen2022maxstyle}, CSDG~\cite{ouyang2022causality}), and various SAM-based models (MedSAM~\cite{ma2024segment}, SAMed~\cite{zhang2023customized}, SAMUS~\cite{lin2023samus}, DeSAM~\cite{gao2024desam}). As detailed in Table~\ref{tab1}, our method consistently outperforms these competing approaches. Our proposed KG-SAM demonstrates superior performance across diverse anatomical regions and imaging modalities. In the abdominal segmentation tasks, it achieves high Dice scores on both MRI (78.05\%) and CT (79.68\%) datasets. For the multi-site prostate segmentation challenge, KG-SAM shows robust performance, achieving an overall Dice score of 82.69\%. These results represent a substantial improvement over existing methods.
To validate the modularity and complementary nature of our approach, we integrated our knowledge-guided framework with the strong DeSAM baseline, creating a model denoted as KG-De-SAM. This integration yields further performance gains across all metrics, culminating in an overall Dice score of 84.52\%. This result corresponds to a significant improvement of +1.83\% over KG-SAM and +2.27\% over DeSAM, confirming that our knowledge-guided mechanism can effectively enhance other advanced SAM-based frameworks. The performance gains are particularly pronounced in challenging cases characterized by ambiguous boundaries or domain shifts. For instance, in abdominal CT segmentation, the Dice score improves from 79.68\% with KG-SAM to 81.68\% with KG-De-SAM. This consistent pattern of enhancement validates both the standalone effectiveness of our framework and its value as a complementary module.

\subsection{Ablation Study}
We conducted a series of ablation experiments to systematically quantify the contribution of each core component in our KG-SAM framework: the Conditional Random Field (CRF), the Knowledge Graph (KG), and the Uncertainty-Aware Fusion (UF). The results, detailed in Table~\ref{tab:ablation}, demonstrate a clear synergistic effect. We evaluate the impact of adding each component individually. Incorporating the CRF module (+CRF) to refine boundaries boosts performance to 76.86\%. Adding the knowledge graph (+KG) to enforce anatomical priors yields a similar gain and the uncertainty fusion mechanism (+UF) also provides a notable improvement. Our full model achieves a final Dice score of 82.69\% on the prostate dataset, a substantial improvement of +10.60\% over the baseline. To further validate the necessity of each component, we evaluated versions of the full model with one module removed. Removing the UF module (w/o UF) results in a score of 79.96\%, while removing the KG (w/o KG) or CRF (w/o CRF) leads to scores of 78.57\% and 79.13\%, respectively. The significant performance drop in each case confirms that all three components are integral to the framework's success and that they work together in a complementary.

\section{Conclusion}
In this paper, we presented KG-SAM, a knowledge-guided framework that enhances the Segment Anything Model for the specific challenges of medical image segmentation. By integrating a knowledge graph, a CRF, and an uncertainty fusion module, KG-SAM effectively enforces anatomical consistency and refines segmentation boundaries. Our extensive experiments validate the superiority of KG-SAM, which achieves an 82.69\% Dice score on a challenging multi-site prostate dataset. The comprehensive ablation studies confirmed the efficacy of our design, with the synergistic combination of components yielding a substantial +10.60\% performance gain over the baseline. Furthermore, we demonstrated the modularity of our approach by integrating it with DeSAM, proving that our framework can serve as a powerful, complementary module for the broader family of SAM-based models. Future work will focus on expanding the knowledge graph to encompass a wider range of anatomical structures and exploring dynamic graph construction methods for patient-specific data.


\vfill\pagebreak

\bibliographystyle{IEEEbib}
\bibliography{refs}

@article{gao2023anatomy,
  title={An anatomy-aware framework for automatic segmentation of parotid tumor from multimodal MRI},
  author={Gao, Yifan and Dai, Yin and Liu, Fayu and Chen, Weibing and Shi, Lifu},
  journal={Computers in Biology and Medicine},
  volume={161},
  pages={107000},
  year={2023},
  publisher={Elsevier}
}

@article{Kirillov2023SegmentA,
  title={Segment Anything},
  author={Alexander Kirillov and Eric Mintun and Nikhila Ravi and Hanzi Mao and Chlo{\'e} Rolland and Laura Gustafson and Tete Xiao and Spencer Whitehead and Alexander C. Berg and Wan-Yen Lo and Piotr Doll{\'a}r and Ross B. Girshick},
  journal={2023 IEEE/CVF International Conference on Computer Vision (ICCV)},
  year={2023},
  pages={3992-4003},
}

@article{ma2024segment,
  title={Segment anything in medical images},
  author={Ma, Jun and He, Yuting and Li, Feifei and Han, Lin and You, Chenyu and Wang, Bo},
  journal={Nature Communications},
  volume={15},
  number={1},
  pages={654},
  year={2024},
  publisher={Nature Publishing Group UK London}
}

@article{zhang2023customized,
  title={Customized segment anything model for medical image segmentation},
  author={Zhang, Kaidong and Liu, Dong},
  journal={arXiv preprint arXiv:2304.13785},
  year={2023}
}

@inproceedings{li2024polyp,
  title={Polyp-sam: Transfer sam for polyp segmentation},
  author={Li, Yuheng and Hu, Mingzhe and Yang, Xiaofeng},
  booktitle={Medical Imaging 2024: Computer-Aided Diagnosis},
  volume={12927},
  pages={759--765},
  year={2024},
  organization={SPIE}
}

@inproceedings{gao2024desam,
  title={Desam: Decoupled segment anything model for generalizable medical image segmentation},
  author={Gao, Yifan and Xia, Wei and Hu, Dingdu and Wang, Wenkui and Gao, Xin},
  booktitle={International Conference on Medical Image Computing and Computer-Assisted Intervention},
  pages={509--519},
  year={2024},
  organization={Springer}
}

@article{Zhang2021UncertaintyGuidedMC,
  title={Uncertainty-Guided Mutual Consistency Learning for Semi-Supervised Medical Image Segmentation},
  author={Yichi Zhang and Qingcheng Liao and Rushi Jiao and Jicong Zhang},
  journal={Artificial intelligence in medicine},
  year={2021},
  volume={138},
  pages={
          102476
        },
}

@article{wang2022conditional,
  title={A conditional random field based feature learning framework for battery capacity prediction},
  author={Wang, Hai-Kun and Zhang, Yang and Huang, Mohong},
  journal={Scientific Reports},
  volume={12},
  number={1},
  pages={13221},
  year={2022},
  publisher={Nature Publishing Group UK London}
}

@article{zhang2024segment,
  title={Segment anything model for medical image segmentation: Current applications and future directions},
  author={Zhang, Yichi and Shen, Zhenrong and Jiao, Rushi},
  journal={Computers in Biology and Medicine},
  pages={108238},
  year={2024},
  publisher={Elsevier}
}

@article{lei2025medlsam,
  title={MedLSAM: Localize and segment anything model for 3D CT images},
  author={Lei, Wenhui and Xu, Wei and Li, Kang and Zhang, Xiaofan and Zhang, Shaoting},
  journal={Medical Image Analysis},
  volume={99},
  pages={103370},
  year={2025},
  publisher={Elsevier}
}

@article{mehrtash2020confidence,
  title={Confidence calibration and predictive uncertainty estimation for deep medical image segmentation},
  author={Mehrtash, Alireza and Wells, William M and Tempany, Clare M and Abolmaesumi, Purang and Kapur, Tina},
  journal={IEEE transactions on medical imaging},
  volume={39},
  number={12},
  pages={3868--3878},
  year={2020},
  publisher={IEEE}
}

@article{xu2022image,
  title={Image segmentation based on higher-order MRF model with multi-node topological overlap measure},
  author={Xu, Sheng-Jun and Zhou, Ying-Xi and Meng, Yue-Bo and Liu, Guang-Hui and Shi, Ya},
  journal={Acta Autom. Sin},
  volume={48},
  pages={1353--1369},
  year={2022}
}

@article{isensee2018nnu,
  title={nnu-net: Self-adapting framework for u-net-based medical image segmentation},
  author={Isensee, Fabian and Petersen, Jens and Klein, Andre and Zimmerer, David and Jaeger, Paul F and Kohl, Simon and Wasserthal, Jakob and Koehler, Gregor and Norajitra, Tobias and Wirkert, Sebastian and others},
  journal={arXiv preprint arXiv:1809.10486},
  year={2018}
}

@inproceedings{chen2020realistic,
  title={Realistic adversarial data augmentation for MR image segmentation},
  author={Chen, Chen and Qin, Chen and Qiu, Huaqi and Ouyang, Cheng and Wang, Shuo and Chen, Liang and Tarroni, Giacomo and Bai, Wenjia and Rueckert, Daniel},
  booktitle={Medical Image Computing and Computer Assisted Intervention--MICCAI 2020: 23rd International Conference, Lima, Peru, October 4--8, 2020, Proceedings, Part I 23},
  pages={667--677},
  year={2020},
  organization={Springer}
}

@article{xu2020robust,
  title={Robust and generalizable visual representation learning via random convolutions},
  author={Xu, Zhenlin and Liu, Deyi and Yang, Junlin and Raffel, Colin and Niethammer, Marc},
  journal={arXiv preprint arXiv:2007.13003},
  year={2020}
}

@inproceedings{chen2022maxstyle,
  title={Maxstyle: Adversarial style composition for robust medical image segmentation},
  author={Chen, Chen and Li, Zeju and Ouyang, Cheng and Sinclair, Matthew and Bai, Wenjia and Rueckert, Daniel},
  booktitle={International Conference on Medical Image Computing and Computer-Assisted Intervention},
  pages={151--161},
  year={2022},
  organization={Springer}
}

@article{ouyang2022causality,
  title={Causality-inspired single-source domain generalization for medical image segmentation},
  author={Ouyang, Cheng and Chen, Chen and Li, Surui and Li, Zeju and Qin, Chen and Bai, Wenjia and Rueckert, Daniel},
  journal={IEEE Transactions on Medical Imaging},
  volume={42},
  number={4},
  pages={1095--1106},
  year={2022},
  publisher={IEEE}
}

@article{lin2023samus,
  title={Samus: Adapting segment anything model for clinically-friendly and generalizable ultrasound image segmentation},
  author={Lin, Xian and Xiang, Yangyang and Zhang, Li and Yang, Xin and Yan, Zengqiang and Yu, Li},
  journal={arXiv preprint arXiv:2309.06824},
  volume={4},
  number={11},
  year={2023}
}

@inproceedings{landman2015miccai,
  title={Miccai multi-atlas labeling beyond the cranial vault--workshop and challenge},
  author={Landman, Bennett and Xu, Zhoubing and Igelsias, J and Styner, Martin and Langerak, T and Klein, Arno},
  booktitle={Proc. MICCAI Multi-Atlas Labeling Beyond Cranial Vault—Workshop Challenge},
  volume={5},
  pages={12},
  year={2015}
}

@article{ji2022amos,
  title={Amos: A large-scale abdominal multi-organ benchmark for versatile medical image segmentation},
  author={Ji, Yuanfeng and Bai, Haotian and Ge, Chongjian and Yang, Jie and Zhu, Ye and Zhang, Ruimao and Li, Zhen and Zhanng, Lingyan and Ma, Wanling and Wan, Xiang and others},
  journal={Advances in neural information processing systems},
  volume={35},
  pages={36722--36732},
  year={2022}
}

@misc{bloch2015nci,
  author = {Bloch, N. and Madabhushi, A. and Huisman, H. and Freymann, J. and Kirby, J. and Grauer, M. and Enquobahrie, A. and Jaffe, C. and Clarke, L. and Farahani, K.},
  title = {NCI-ISBI 2013 Challenge: Automated Segmentation of Prostate Structures},
  year = {2015},
  publisher = {The Cancer Imaging Archive},
  doi = {10.7937/K9/TCIA.2015.zF0vlOPv}
}

@article{lemaitre2015computer,
  title={Computer-aided detection and diagnosis for prostate cancer based on mono and multi-parametric MRI: a review},
  author={Lema{\^\i}tre, Guillaume and Mart{\'\i}, Robert and Freixenet, Jordi and Vilanova, Joan C and Walker, Paul M and Meriaudeau, Fabrice},
  journal={Computers in biology and medicine},
  volume={60},
  pages={8--31},
  year={2015},
  publisher={Elsevier}
}

@article{litjens2014evaluation,
  title={Evaluation of prostate segmentation algorithms for MRI: the PROMISE12 challenge},
  author={Litjens, Geert and Toth, Robert and Van De Ven, Wendy and Hoeks, Caroline and Kerkstra, Sjoerd and Van Ginneken, Bram and Vincent, Graham and Guillard, Gwenael and Birbeck, Neil and Zhang, Jindang and others},
  journal={Medical image analysis},
  volume={18},
  number={2},
  pages={359--373},
  year={2014},
  publisher={Elsevier}
}

@article{li2025dual,
  title={Dual branch segment anything model-transformer fusion network for accurate breast ultrasound image segmentation},
  author={Li, Yu and Huang, Jin and Zhang, Yimin and Deng, Jingwen and Zhang, Jingwen and Dong, Lan and Wang, Du and Mei, Liye and Lei, Cheng},
  journal={Medical Physics},
  volume={52},
  number={6},
  pages={4108--4119},
  year={2025},
  publisher={Wiley Online Library}
}

@article{li2025sfmdiffusion,
  title={SfMDiffusion: self-supervised monocular depth estimation in endoscopy based on diffusion models},
  author={Li, Yu and Chang, Da and Luo, Die and Huang, Jin and Dong, Lan and Wang, Du and Mei, Liye and Lei, Cheng},
  journal={International Journal of Computer Assisted Radiology and Surgery},
  pages={1--9},
  year={2025},
  publisher={Springer}
}

\end{document}